\documentclass[letterpaper,10pt,conference]{ieeeconf}

\IEEEoverridecommandlockouts
\usepackage{amsmath,amssymb,amsfonts}
\usepackage{bm}
\usepackage{graphicx}
\usepackage{xcolor}
\usepackage{multirow}
\usepackage{arydshln}
\usepackage{threeparttable} 
\usepackage{hyperref}

\newcommand{\SO}{\mathbb{SO}(3)}
\newcommand{\SE}{\mathbb{SE}(3)}
\newcommand{\SEK}[1]{\mathbb{SE}_{#1}(3)}
\newcommand{\Exp}{\operatorname{Exp}}
\newcommand{\Log}{\operatorname{Log}}

\newcommand{\N}{\mathcal{N}}
\newcommand{\best}[1]{{\color{red!70!black}\bfseries #1}}
\newcommand{\bestmath}[1]{{\color{red!70!black}\mathbf{#1}}}

\title{Consistent Distributed Invariant EKF SLAM with Point and Object Features}
\author{Mengya Xu, Xianjia Yu, Ha Sier, Shoudong Huang, Tomi Westerlund
}

\begin{document}
\maketitle

\begin{abstract}

Multi-robot SLAM allows robots to improve their estimates by exchanging map information, but repeated exchanges create unknown correlations between local estimators.
Covariance intersection (CI) can fuse estimates without tracking these correlations.
In right-invariant extended Kalman filter (RI-EKF) SLAM, however, point features obtain their invariant errors through the robot rotation, so a neighbor's map estimate is expressed in its own error coordinates and is correlated with its pose, and cannot be fused by CI directly.
This paper presents Joint-CI, a unified fusion method for distributed RI-EKF SLAM with point, object, and mixed point--object maps.
Using a neighboring robot's local posterior and an inter-robot observation, a joint estimate of the receiving robot's pose and the common features is constructed in the receiving robot's error coordinates, retaining the robot--feature and feature--feature correlations.
Partial-state CI then fuses this estimate with the receiving robot's local posterior.
We show that the unified formulation and the fusion preserve the six unobservable directions associated with global rotation and translation, and that communication and fusion preserve first-order covariance consistency given consistent local posteriors.
Monte Carlo simulations and a real-world dataset-derived experiment show that Joint-CI improves estimation accuracy while maintaining consistent, mildly conservative uncertainty.
The code is available at
\url{https://anonymous.4open.science/r/Joint-CI-for-Distributed-RI-EKF-SLAM-E14B/}.
\end{abstract}

\section{Introduction}
Multi-robot SLAM improves coverage and estimation accuracy by allowing robots to exchange locally acquired map information. 
Representative systems such as DDF-SAM~\cite{6631323}, Kimera-Multi~\cite{9686955}, and Swarm-SLAM~\cite{10321649} employ distributed or decentralized nonlinear optimization over factor-graph or pose-graph representations. 
These approaches provide powerful batch or incremental optimization capabilities, but repeatedly solving a growing optimization problem can impose computational and latency costs on lightweight robotic platforms. 
Filtering-based methods provide a complementary solution by recursively updating the current posterior as new measurements and communicated estimates arrive.

Point features are generic and efficient to detect and match, and have therefore been widely used in both single-robot and multi-robot SLAM.
Advances in machine learning have also increased interest in high-level features such as objects, which provide persistent semantic identities and strong geometric constraints \cite{GALVEZLOPEZ2016435,epoch, 8490953,8440105}.
Practical systems have further demonstrated the value of mixed map representations, in which low-level point features and high-level object features are maintained together \cite{8708251,9665919,10611684}.
For filtering-based SLAM, estimation consistency is essential because underestimated uncertainty can degrade both state estimation and uncertainty-aware downstream decisions~\cite{9561951}.
Conventional EKF SLAM can become inconsistent because linearization at the estimated states may introduce spurious information along unobservable directions~\cite{CASTELLANOS2004716,4339532,4543252}.
Right-invariant EKF (RI-EKF) formulations address this issue by defining the estimation error through the symmetry of the state space, which preserves the appropriate unobservable directions.
They have been developed for both point-feature SLAM~\cite{barrau2016ekfslam,7812660} and object-feature SLAM~\cite{9667208}.

Distributed filtering introduces an additional difficulty because previous information exchange creates generally unknown cross-correlations between local estimators. 
Covariance intersection (CI) \cite{609105} provides a conservative fusion rule without explicitly tracking these correlations and has been applied to SLAM~\cite{JULIER20073} and decentralized cooperative localization~\cite{6696534,9531527}. 
Partial-state CI extends this idea to estimates containing different or overlapping subsets of a common state~\cite{ajgl2014partialci,7528001,trumpf2025partialci}.
Distributed invariant filters have also applied CI to pose estimation on Lie groups~\cite{10232373,10886693}, without estimating a shared map. 
For object-level multi-robot SLAM, Li \textit{et al.}~\cite{li2025objectslam} use per-object invariant errors, representing each object as an independent $\SE$ component, so that neighboring robot poses and common object estimates can be combined separately.

Despite these advances, fusing map information between RI-EKF SLAM estimators remains nontrivial.
Exchanging only pose estimates discards the information contained in the shared map, and the per-object invariant errors in~\cite{li2025objectslam} rely on object orientations and therefore do not directly cover point features.
In RI-EKF SLAM, point features, which carry no orientation of their own, obtain their invariant errors through the robot rotation~\cite{barrau2016ekfslam,7812660}, and the object formulation in~\cite{9667208} adopts the same coupling, allowing point, object, and mixed maps to share one error representation.
In this coupled formulation, exchanging map features raises two issues.
First, each robot's feature errors are defined relative to its own rotation error, so a neighbor's feature covariance cannot be combined directly with the receiving robot's posterior.
Second, a neighbor's pose and feature errors are correlated through its own SLAM estimation, and in the coupled formulation they additionally share the same rotation error. 
A communicated covariance that ignores these correlations is not guaranteed to be consistent.

This paper presents Joint-CI, a single construction that addresses both issues for distributed RI-EKF SLAM with point, object, and mixed point--object maps.
Joint-CI re-expresses the communicated robot pose and common features in the receiving filter's error coordinates as one correlated partial state, derives its complete first-order covariance from the neighboring posterior and the inter-robot observation, and then applies standard partial-state CI.
The main contributions are:
1) We derive a unified construction of the communicated robot--feature partial state and its complete first-order covariance in error coordinates compatible with the receiving filter, for point, object, and mixed maps.
2) We show that the unified local RI-EKF formulation and the subsequent fusion preserve the global gauge directions. Given consistent local posteriors, we further establish first-order covariance consistency of the communicated estimate and subsequent CI fusion.  
3) We evaluate estimation accuracy, statistical consistency, communication and computational costs, and operation in multi-neighbor networks through Monte Carlo simulations and a real-world dataset-derived experiment.

\section{Problem formulation}
We consider \(K\) robots, \(N_p\) point features, and \(N_o\) object features, indexed by \(k\), \(a\), and \(b\), respectively.
Robot \(k\) has orientation \(\mathbf R_k\in\SO\) and position \(\mathbf x_k\in\mathbb R^3\).
Point feature \(a\) has position \(\mathbf q^a\in\mathbb R^3\), while object feature \(b\) has orientation \(\mathbf R^{f_b}\in\SO\) and position \(\mathbf f^b\in\mathbb R^3\).
All robot and feature states are expressed in a common global frame.

\subsubsection{State Space}
Robot $k$ maintains a local map with point-feature index set $\mathcal I_k^p$ and object-feature index set $\mathcal I_k^o$.
Its state can be written as
\begin{equation}
\begin{split}
    \mathbf X_k=(&\mathbf R_k,
      \{\mathbf R^{f_b}_k\}_{b\in\mathcal I_k^o},
      \mathbf x_k,
      \{\mathbf q^a_k\}_{a\in\mathcal I_k^p},
      \{\mathbf f^b_k\}_{b\in\mathcal I_k^o}).
\end{split}
\label{eq:local-state}
\end{equation}

For compactness, feature index ranges are omitted in subsequent equations.
Let $m_p=|\mathcal I_k^p|$ and $m_o=|\mathcal I_k^o|$ denote the numbers of point and object features in robot $k$'s local map.
We define the coupled state group $\mathcal G_{m_p,m_o}$, which is isomorphic to  $\SEK{1+m_p+m_o}\times\SO^{m_o}$, where $\SEK{m}$ contains one rotation and $m$ three-dimensional vectors~\cite{7812660,barrau2016ekfslam}.
The local state satisfies $\mathbf X_k\in\mathcal G_{m_p,m_o}$.
When the numbers of point and object features are clear from context, $\mathcal G$ denotes the corresponding group $\mathcal G_{m_p,m_o}$.

\subsubsection{Process model}
Let $n$ denote the discrete time index.
Following the first-order integration scheme commonly used in RI-EKF SLAM \cite{7812660,9667208}, the process model is
\begin{equation}
\begin{split}
    \mathbf R_{k,n+1}
      &=\mathbf R_{k,n}\Exp(\mathbf w^R_{k,n})
        \mathbf R^u_{k,n},\\
    \mathbf x_{k,n+1}
      &=\mathbf x_{k,n}+\mathbf R_{k,n}
        (\mathbf v_{k,n}+\mathbf w^v_{k,n}),
\end{split}
\label{eq:process}
\end{equation}
where $(\mathbf R^u_{k,n}, \mathbf v_{k,n})$ is the odometry, $\mathbf w_{k,n}=[(\mathbf w^R_{k,n})^T, (\mathbf w^v_{k,n})^T]^T\sim\N(\mathbf0,\bm\Phi_{k,n})$ is the corresponding process noise, and $\Exp:\mathbb R^3\rightarrow\SO$ denotes the exponential map, with local inverse $\Log$.
Features are static.

\subsubsection{Observation models}
Robot $k$ may obtain three types of observations, including point feature observations, object feature pose observations, and inter-robot pose observations.
Here, the time index is omitted from the observation models for readability.

The observation of point feature $a$ by robot $k$ is
\begin{equation}
    \mathbf z^q_{ka}
    =\mathbf R_k^T(\mathbf q^a-\mathbf x_k)+\bm\nu^q_{ka},
\label{eq:point-observation}
\end{equation}
where $\bm\nu^q_{ka}\sim\N(\mathbf0,\bm\Psi^q_{ka})$ is the point observation noise.

The observation of object feature $b$ by robot $k$ is
\begin{equation}
\begin{split}
    \mathbf R^z_{kb}
       &=\Exp(\bm\nu^R_{kb})\mathbf R_k^T\mathbf R^{f_b},\\
    \mathbf z^f_{kb}
       &=\mathbf R_k^T(\mathbf f^b-\mathbf x_k)+\bm\nu^f_{kb},
\end{split}
\label{eq:object-observation}
\end{equation}
where $\bm\nu^o_{kb}=[(\bm\nu^R_{kb})^T,(\bm\nu^f_{kb})^T]^T\sim\N(\mathbf0,\bm\Psi^o_{kb})$ is the object pose observation noise.

Let $l\neq k$ denote another robot.
When robot $k$ observes robot $l$, the inter-robot observation is
\begin{equation}
\begin{split}
    \mathbf R^z_{kl}
       &=\Exp(\bm\nu^R_{kl})\mathbf R_k^T\mathbf R_l,\\
    \mathbf z^x_{kl}
       &=\mathbf R_k^T(\mathbf x_l-\mathbf x_k)+\bm\nu^x_{kl},
\end{split}
\label{eq:robot-observation}
\end{equation}
where $\bm\nu_{kl}=[(\bm\nu^R_{kl})^T,(\bm\nu^x_{kl})^T]^T\sim\N(\mathbf0,\bm\Psi_{kl})$ is the inter-robot pose observation noise.

\subsubsection{Problem of Interest}
Each robot maintains a local RI-EKF posterior over its own pose and locally observed features.
Robots communicate over a time-varying graph, and $\mathcal V_k$ denotes the set of robots that can communicate with and observe robot $k$ at the current time.
For each $l\in\mathcal V_k$, the inter-robot observation allows robot $l$ to estimate robot $k$'s pose, while its local posterior provides estimates of their common features.
The objective is to construct from each neighbor a joint estimate of robot $k$'s pose and the common features, with a covariance expressed in invariant error coordinates compatible with robot $k$'s local posterior, and to fuse all such estimates with robot $k$'s local posterior without tracking the unknown cross-correlations among the local estimators.
We assume that all robots are initialized in a common but arbitrary global frame with consistent initial covariances, and that correspondences between features observed by different robots are available.
All process and observation noises are assumed mutually independent across robots, features, and time.
Each current noise realization is also assumed independent of the prior state-estimation errors.

\section{Distributed RI-EKF SLAM}

\subsection{Unified local RI-EKF formulation}
The local estimator follows the standard RI-EKF propagation, measurement update, and feature initialization procedures established for point-feature SLAM \cite{barrau2016ekfslam,7812660} and object-feature SLAM \cite{9667208}.
Both formulations use the robot rotation perturbation in the error definition of every position component, while object features additionally contain an orientation error. 
This shared structure yields the unified error defined below.

For any $\mathbf X_1,\mathbf X_2\in\mathcal G_{m_p,m_o}$, the operator $\oplus$ is defined by
\begin{equation}
	\begin{aligned}
		\mathbf X_1\oplus\mathbf X_2
		&=\bigl(
		\mathbf R_1\mathbf R_2,
		\{\mathbf R^{f_b}_1\mathbf R^{f_b}_2\},\\
		&\mathbf R_1\mathbf x_2+\mathbf x_1,
		\{\mathbf R_1\mathbf q^a_2+\mathbf q^a_1\},
		\{\mathbf R_1\mathbf f^b_2+\mathbf f^b_1\}
		\bigr).
	\end{aligned}
\end{equation}

The associated minus operation is defined as $\mathbf X_1\ominus\mathbf X_2=\mathbf X_1\oplus\mathbf X_2^{-1}$, where $\mathbf X_2^{-1}$ is the group inverse.

Identifying the Lie algebra with its vector coordinates, the error $\mathbf e\in\mathbb R^{6+3m_p+6m_o}$ is
\begin{equation}
\begin{split}
\mathbf e^T=
\bigl[&
(\mathbf e^R)^T,
\{(\mathbf e^{R^{f_b}})^T\},
(\mathbf e^x)^T,
\{(\mathbf e^{q^a})^T\},
\{(\mathbf e^{f^b})^T\}
\bigr],
\end{split}
\end{equation}
where $\mathbf e^R,\mathbf e^x, \mathbf e^{q^a}, \mathbf e^{R^{f_b}},\mathbf e^{f^b} \in\mathbb R^3$ for all applicable $a$ and $b$.

The exponential map on this Lie group is defined by
\begin{equation}
\begin{split}
\Exp_{\mathcal G}(\mathbf e)=(&\Exp(\mathbf e^R),
\{\Exp(\mathbf e^{R^{f_b}})\},\\
&J(\mathbf e^R)\mathbf e^x,
\{J(\mathbf e^R)\mathbf e^{q^a}\},\{J(\mathbf e^R)\mathbf e^{f^b}\}), 
\end{split}
\label{eq:group-exp}
\end{equation}
where $J(\cdot)$ denotes the left Jacobian of $\SO$ under the convention of \cite{9667208}.

The true state $\mathbf X$, its estimate $\hat{\mathbf X}$, and the associated right-invariant error $\mathbf e$ satisfy
\begin{equation}
    \mathbf X=\Exp_{\mathcal G}(\mathbf e)\oplus\hat{\mathbf X}.
\label{eq:retraction}
\end{equation}
Each robot~$k$ maintains the estimate $\hat{\mathbf X}_k$ and the covariance $\mathbf P_k$ of its error $\mathbf e_k$.
Time indices are added when needed.

According to the process model \eqref{eq:process}, the propagated invariant error of robot $k$, $\mathbf e_{k,n+1|n}$, satisfies
\begin{equation}
	\begin{split}
		\mathbf e_{k,n+1|n} 
		\simeq \mathbf F_{k,n}\mathbf e_{k,n}
		+\mathbf G_{k,n}\mathbf w_{k,n}.
	\end{split}
	\label{eq:mixed-error-propagation}
\end{equation}
The corresponding first-order Jacobians are
\begin{equation}
	\begin{aligned}
		\mathbf F_{k,n} &=\mathbf I_{6+3m_p+6m_o},\\
		\mathbf G_{k,n} &=
		\begin{bmatrix}
			\hat{\mathbf R}_{k,n}&\mathbf0\\
			\mathbf0&\mathbf0\\
			[\hat{\mathbf x}_{k,n+1|n}]_\times\hat{\mathbf R}_{k,n}
			&\hat{\mathbf R}_{k,n}\\
			\operatorname{col}_{a\in\mathcal I_k^p}\left([\hat{\mathbf q}_{k,n}^a]_\times\hat{\mathbf R}_{k,n}\right)&\mathbf0\\
			\operatorname{col}_{b\in\mathcal I_k^o}\left([\hat{\mathbf f}_{k,n}^b]_\times\hat{\mathbf R}_{k,n}\right)&\mathbf0
		\end{bmatrix}
	\end{aligned}
\label{eq:mixed-propagation-jacobians}
\end{equation}
where $[\cdot]_\times$ denotes the skew-symmetric matrix of a vector and $\operatorname{col}_{i\in\mathcal I}(\mathbf M_i)$ stacks the matrices $\mathbf M_i$ vertically according to the state ordering.

At a measurement update, the time indices are omitted for readability.
For point feature $a$ and object feature $b$, define the residuals
\begin{equation}
\begin{split}
\mathbf r^q_{ka}
&=\mathbf z^q_{ka}
-\hat{\mathbf R}_k^T(\hat{\mathbf q}^a_k-\hat{\mathbf x}_k),\\
\mathbf r^o_{kb}
&=\begin{bmatrix}
\Log\!\left(\mathbf R^z_{kb}
(\hat{\mathbf R}_k^T\hat{\mathbf R}^{f_b}_k)^T\right)\\
\mathbf z^f_{kb}
-\hat{\mathbf R}_k^T(\hat{\mathbf f}^b_k-\hat{\mathbf x}_k)
\end{bmatrix}.
\end{split}
\label{eq:mixed-residuals}
\end{equation}
Their first-order models are
\begin{equation}
\mathbf r^q_{ka}\simeq\mathbf H^q_{ka}\mathbf e_k
+\bm\nu^q_{ka},
\qquad
\mathbf r^o_{kb}\simeq\mathbf H^o_{kb}\mathbf e_k
+\bm\nu^o_{kb}.
\label{eq:mixed-residual-linearization}
\end{equation}
Let $\mathbf S_a^p\in\mathbb R^{3\times3m_p}$ and $\mathbf S_b^o\in\mathbb R^{3\times3m_o}$ select point $a$ and object $b$ according to the state ordering.
The corresponding Jacobians are
\begin{equation}
\begin{split}
\mathbf H^q_{ka}
&=\begin{bmatrix}
\mathbf0&\mathbf0&-\hat{\mathbf R}_k^T
&\hat{\mathbf R}_k^T\mathbf S_a^p&\mathbf0
\end{bmatrix},\\
\mathbf H^o_{kb}
&=\begin{bmatrix}
-\hat{\mathbf R}_k^T&\hat{\mathbf R}_k^T\mathbf S_b^o
&\mathbf0&\mathbf0&\mathbf0\\
\mathbf0&\mathbf0&-\hat{\mathbf R}_k^T
&\mathbf0&\hat{\mathbf R}_k^T\mathbf S_b^o
\end{bmatrix}.
\end{split}
\label{eq:mixed-observation-jacobians}
\end{equation}
The residuals and Jacobians of all currently observed features are stacked for the measurement update.

When a feature is first observed, its estimate is initialized using the inverse observation model:
\begin{equation}
\begin{aligned}
\hat{\mathbf q}^a_k
&=\hat{\mathbf x}_k+\hat{\mathbf R}_k\mathbf z^q_{ka},\\
(\hat{\mathbf R}^{f_b}_k, \hat{\mathbf f}^b_k)
&=(\hat{\mathbf R}_k\mathbf R^z_{kb},\hat{\mathbf x}_k+\hat{\mathbf R}_k\mathbf z^f_{kb}).
\end{aligned}
\label{eq:mixed-feature-initialization}
\end{equation}
Under the error convention in \eqref{eq:retraction}, the corresponding first-order errors satisfy
\begin{equation}
\begin{aligned}
\mathbf e^{q^a}
&\simeq\mathbf e^x-\hat{\mathbf R}_k\bm\nu^q_{ka},\\
\begin{bmatrix}
\mathbf e^{R^{f_b}}\\
\mathbf e^{f^b}
\end{bmatrix}
&\simeq
\begin{bmatrix}
\mathbf e^R\\
\mathbf e^x
\end{bmatrix}
-\operatorname{blkdiag}(\hat{\mathbf R}_k,\hat{\mathbf R}_k)
\bm\nu^o_{kb},
\end{aligned}
\label{eq:mixed-initialization-errors}
\end{equation}
where $\operatorname{blkdiag}(\cdot)$ forms a block-diagonal matrix from its arguments.

For $s\in\{q,o\}$, let $\mathbf e_{\mathrm{new}}^s$, $\bm\nu_k^s$, and $\bm\Psi_k^s$ denote the corresponding new-feature error, observation noise, and noise covariance, respectively, with the feature index $a$ or $b$ omitted.
The relations in \eqref{eq:mixed-initialization-errors} can be written as
\begin{equation}
\mathbf e_{\mathrm{new}}^s
\simeq
\mathbf L_k^s\mathbf e_k+\mathbf B_k^s\bm\nu_k^s,
\end{equation}
where $\mathbf L_k^s$ and $\mathbf B_k^s$ are the corresponding state-error and observation-noise Jacobians.
Let $\bm\Pi_k^s$ be the permutation matrix that maps the appended ordering $[\mathbf e_k^T,(\mathbf e_{\mathrm{new}}^s)^T]^T$ to the mixed-state ordering.
The augmented covariance is then
\begin{equation}
\mathbf P_{k,\mathrm{aug}}^s
=
\bm\Pi_k^s
\begin{bmatrix}
\mathbf I&\mathbf0\\
\mathbf L_k^s&\mathbf B_k^s
\end{bmatrix}
\begin{bmatrix}
\mathbf P_k&\mathbf0\\
\mathbf0&\bm\Psi_k^s
\end{bmatrix}
\begin{bmatrix}
\mathbf I&\mathbf0\\
\mathbf L_k^s&\mathbf B_k^s
\end{bmatrix}^{T}
(\bm\Pi_k^s)^T.
\label{eq:mixed-covariance-augmentation}
\end{equation}

The mixed-state formulation includes point-only and object-only RI-EKF SLAM as the special cases $m_o=0$ and $m_p=0$, respectively.

\subsection{Joint communicated state in compatible coordinates}
\label{sec:communicated-state}
A communicated error representation is compatible with robot \(k\)’s local state when the corresponding components use the same invariant retraction, tangent coordinates, and ordering after selection.
The resulting residual and covariance can then be used directly in partial-state CI without further coordinate transformation.

Consider a neighbor $l\in\mathcal V_k$.
Let $\mathcal C^p_{kl}=\mathcal I_k^p\cap\mathcal I_l^p$ and $\mathcal C^o_{kl}=\mathcal I_k^o\cap\mathcal I_l^o$ denote their common point and object features.
To transmit this information for fusion with robot \(k\)’s local posterior, robot \(l\) must construct a joint estimate and express its covariance in error coordinates compatible with robot \(k\)’s local state.

Using \eqref{eq:robot-observation}, robot $l$ first constructs a virtual pose of robot $k$ as
\begin{equation}
\begin{aligned}
(\hat{\mathbf R}_{l\rightarrow k},\hat{\mathbf x}_{l\rightarrow k})
=\bigl(\hat{\mathbf R}_l\mathbf R^z_{lk},
\hat{\mathbf x}_l+\hat{\mathbf R}_l\mathbf z^x_{lk}\bigr).
\end{aligned}
\label{eq:virtual-pose}
\end{equation}
The communicated joint estimate is constructed from this virtual pose and robot $l$'s estimates of the common features:
\begin{equation}
\begin{split}
\hat{\mathbf Y}_{l\rightarrow k}=(&
\hat{\mathbf R}_{l\rightarrow k},
\{\hat{\mathbf R}^{f_b}_l\}_{b\in\mathcal C^o_{kl}},\\
&\hat{\mathbf x}_{l\rightarrow k},
\{\hat{\mathbf q}^a_l\}_{a\in\mathcal C^p_{kl}},
\{\hat{\mathbf f}^b_l\}_{b\in\mathcal C^o_{kl}}).
\end{split}
\label{eq:communicated-state}
\end{equation}
Let $\mathbf Y_{l\rightarrow k}$ denote the true counterpart of \eqref{eq:communicated-state}.
Its error is defined using the same invariant error representation as robot \(k\)’s corresponding partial state, with components ordered as
\begin{equation}
\begin{aligned}
	\mathbf e_{l\rightarrow k}
	&=\Log_{\mathcal G}\!\left(
	\mathbf Y_{l\rightarrow k}\ominus
	\hat{\mathbf Y}_{l\rightarrow k}\right)\\
	&=\bigl[
	(\mathbf e_{l\rightarrow k}^{R})^T,
	\{(\mathbf e_{l\rightarrow k}^{R^{f_b}})^T\},
	(\mathbf e_{l\rightarrow k}^{x})^T,\\
	&\hspace{4mm}
	\{(\mathbf e_{l\rightarrow k}^{q^a})^T\},
	\{(\mathbf e_{l\rightarrow k}^{f^b})^T\}
	\bigr]^T,
\end{aligned}
\label{eq:communicated-error-ordering}
\end{equation}
where $\Log_{\mathcal G}$ denotes the local inverse of $\Exp_{\mathcal G}$.

Let $\mathbf e_l^{\mathcal S}$ denote the subvector of robot $l$'s local error associated with its pose and the common features:
\begin{equation}
\begin{split}
(\mathbf e_l^{\mathcal S})^T
=\bigl[&
(\mathbf e_l^R)^T,
\bigl\{(\mathbf e_l^{R^{f_b}})^T\bigr\}_{b\in\mathcal C_{kl}^o},\\
&(\mathbf e_l^x)^T,
\bigl\{(\mathbf e_l^{q^a})^T\bigr\}_{a\in\mathcal C_{kl}^p},
\bigl\{(\mathbf e_l^{f^b})^T\bigr\}_{b\in\mathcal C_{kl}^o}
\bigr].
\end{split}
\label{eq:selected-source-error}
\end{equation}
Its covariance $\mathbf P_l^{\mathcal S}$ is the principal submatrix of robot $l$'s local covariance $\mathbf P_l$ associated with the same state components.

From \eqref{eq:group-exp} and \eqref{eq:retraction}, the rotation error of the robot component acts on every position component.
In robot $l$'s local state, this common rotation error is $\mathbf e_l^R$, whereas in the communicated state it is $\mathbf e_{l\rightarrow k}^{R}$.
The common-feature estimates remain unchanged, but their tangent errors must be re-expressed relative to the rotation error of robot $k$'s virtual pose.
The resulting first-order relations are
\begin{equation}
\begin{aligned}
\mathbf e_{l\rightarrow k}^{R}
&\simeq\mathbf e_l^R-\hat{\mathbf R}_l\bm\nu^R_{lk}, \quad\mathbf e_{l\rightarrow k}^{R^{f_b}} \simeq\mathbf e_l^{R^{f_b}},\\
\mathbf e_{l\rightarrow k}^{x}
&\simeq\mathbf e_l^x-[\hat{\mathbf x}_{l\rightarrow k}]_\times
\hat{\mathbf R}_l\bm\nu^R_{lk}
-\hat{\mathbf R}_l\bm\nu^x_{lk},\\
\mathbf e_{l\rightarrow k}^{y}
&\simeq\mathbf e_l^{y}-[\hat{\mathbf y}]_\times
\hat{\mathbf R}_l\bm\nu^R_{lk},
\end{aligned}
\label{eq:compatible-error-blocks}
\end{equation}
where $(\hat{\mathbf y},\mathbf e_l^y)$ denotes the estimate and local error of any communicated point or object position.
Stacking these relations according to \eqref{eq:communicated-error-ordering} gives $\mathbf e_{l\rightarrow k}\simeq\mathbf e_l^{\mathcal S}+\mathbf D_{l\rightarrow k}\bm\nu_{lk}$, where
\begin{equation}
\begin{split}
\mathbf D_{l\rightarrow k}
&=
\begin{bmatrix}
-\hat{\mathbf R}_l&\mathbf0\\
\mathbf0&\mathbf0\\
-[\hat{\mathbf x}_{l\rightarrow k}]_\times\hat{\mathbf R}_l
&-\hat{\mathbf R}_l\\
-\operatorname{col}_{a\in\mathcal C^p_{kl}}
([\hat{\mathbf q}^a_l]_\times\hat{\mathbf R}_l)&\mathbf0\\
-\operatorname{col}_{b\in\mathcal C^o_{kl}}
([\hat{\mathbf f}^b_l]_\times\hat{\mathbf R}_l)&\mathbf0
\end{bmatrix}.
\end{split}
\label{eq:compatible-error-mapping}
\end{equation}
The two block columns of $\mathbf D_{l\rightarrow k}$ correspond to $\bm\nu^R_{lk}$ and $\bm\nu^x_{lk}$, respectively.
Assuming that the inter-robot observation noise is independent of robot $l$'s local estimation error, the communicated covariance is
\begin{equation}
\mathbf P_{l\rightarrow k}
=\mathbf P_l^{\mathcal S}
+\mathbf D_{l\rightarrow k}\bm\Psi_{lk}
\mathbf D_{l\rightarrow k}^T.
\label{eq:joint-covariance}
\end{equation}

The mapping in \eqref{eq:compatible-error-mapping} changes the robot rotation error shared by the feature-position errors from that of robot $l$ to that of the virtual pose of robot $k$.
Consequently, \(\mathbf P_{l\rightarrow k}\) is expressed in invariant error coordinates compatible with robot \(k\)’s local state.
It retains the robot--feature and feature--feature covariance blocks inherited from $\mathbf P_l$, together with the correlations induced by the inter-robot observation noise.

\subsection{Partial-state covariance intersection}
The communicated estimate contains robot \(k\)’s pose and the common features, while its local posterior may contain additional features.
Since Section~\ref{sec:communicated-state} expresses the corresponding components in a common error space, robot \(k\) fuses the two estimates using standard partial-state CI \cite{trumpf2025partialci}.

Let $\mathbf E_{kl}$ be the binary selection matrix such that $\mathbf e_k^c=\mathbf E_{kl}\mathbf e_k$ contains robot $k$'s pose error and the common-feature errors in the same order as $\mathbf e_{l\rightarrow k}$.
Let $\hat{\mathbf Y}_k^c$ denote the corresponding local partial-state estimate of robot $k$ with the same ordering.
The invariant fusion residual is defined as
\begin{equation}
    \mathbf d_{lk}
    =\Log_{\mathcal G}
       (\hat{\mathbf Y}_{l\rightarrow k}\ominus
        \hat{\mathbf Y}_k^c).
\label{eq:ci-residual}
\end{equation}
Under the first-order error model, this residual satisfies
\begin{equation}
\mathbf d_{lk}\simeq\mathbf E_{kl}\mathbf e_k-\mathbf e_{l\rightarrow k}.
\label{eq:ci-residual-errors}
\end{equation}
Then, the partial-state CI covariance and correction are
\begin{equation}
\begin{split}
    (\mathbf P_k^+)^{-1}
    ={}&\omega\mathbf P_k^{-1}
    +(1-\omega)\mathbf E_{kl}^T
      \mathbf P_{l\rightarrow k}^{-1}\mathbf E_{kl},\\
      \Delta\mathbf e_k
      ={}&(1-\omega)\mathbf P_k^+
      \mathbf E_{kl}^T\mathbf P_{l\rightarrow k}^{-1}
      \mathbf d_{lk},
\end{split}
\label{eq:partial-ci-update}
\end{equation}
where \(\omega\in[0,1]\) is selected by minimizing \(\operatorname{tr}(\mathbf P_k^+)\) over weights for which the fused information matrix is positive definite.
If the communicated state omits local components, this requires $\omega>0$.

The state estimate is updated as $\hat{\mathbf X}_k^+=\Exp_{\mathcal G}(\Delta\mathbf e_k)\oplus\hat{\mathbf X}_k$, and \((\hat{\mathbf X}_k^+,\mathbf P_k^+)\) becomes robot $k$'s local posterior for the subsequent RI-EKF recursion and communication events.

For multiple neighbors, the communicated estimate and covariance are constructed for each \(l\in\mathcal V_k\) as in Section~\ref{sec:communicated-state}, and fused with robot \(k\)’s local posterior using multi-estimate partial-state CI \cite{ajgl2014partialci}:
\begin{equation}
\begin{aligned}
(\mathbf P_k^+)^{-1}
&=\omega_0\mathbf P_k^{-1}
+\sum_{l\in\mathcal V_k}\omega_l\mathbf E_{kl}^T
\mathbf P_{l\rightarrow k}^{-1}\mathbf E_{kl},\\
\Delta\mathbf e_k
&=\mathbf P_k^+
\sum_{l\in\mathcal V_k}\omega_l\mathbf E_{kl}^T
\mathbf P_{l\rightarrow k}^{-1}\mathbf d_{lk}.
\end{aligned}
\label{eq:multi-neighbor-ci}
\end{equation}
The nonnegative weights sum to one and are selected by minimizing \(\operatorname{tr}(\mathbf P_k^+)\) subject to a positive-definite fused information matrix.
For a single neighbor, $\omega_0=\omega$ and $\omega_l=1-\omega$, which recovers \eqref{eq:partial-ci-update}.

\section{Consistency Analysis}

\subsection{Gauge Preservation}

Consider a fixed change of global frame $\mathbf T_g=(\mathbf R_g,\mathbf t_g)\in\SE$, which maps every orientation $\mathbf R$ to $\mathbf R_g\mathbf R$ and every position $\mathbf p$ to $\mathbf R_g\mathbf p+\mathbf t_g$.
Direct substitution into \eqref{eq:process}--\eqref{eq:robot-observation} shows that the process model preserves this transformation and that all observations remain unchanged.
States that differ only by this transformation are therefore indistinguishable from the available motion and observation data.

Let $\boldsymbol\phi,\boldsymbol\rho\in\mathbb R^3$ denote infinitesimal global rotation and translation perturbations, and define $\boldsymbol\xi_g=[\boldsymbol\phi^T,\boldsymbol\rho^T]^T$.
Expressing the globally transformed state through the retraction \eqref{eq:retraction} and retaining first-order terms gives
\(\mathbf e_g=\mathbf N_{m_p,m_o}\boldsymbol\xi_g\), where
\begin{equation}
\mathbf N_{m_p,m_o}=
\begin{bmatrix}
\mathbf I & \mathbf0\\
\mathbf1_{m_o}\otimes\mathbf I & \mathbf0\\
\mathbf0 & \mathbf I\\
\mathbf0 & \mathbf1_{m_p}\otimes\mathbf I\\
\mathbf0 & \mathbf1_{m_o}\otimes\mathbf I
\end{bmatrix}.
\label{eq:mixed-gauge-basis}
\end{equation}
Here, $\mathbf1_m\otimes\mathbf I$ stacks $m$ copies of the three-dimensional identity matrix.
The six columns of $\mathbf N_{m_p,m_o}$ represent the global rotation and translation directions.

Let $\mathbf H_{k,n}$ denote the stacked Jacobian of all feature observations available to robot $k$ at time $n$.
For a fixed local feature set, the mixed-state Jacobians satisfy
\begin{equation}
\mathbf F_{k,n}\mathbf N_{m_p,m_o}=\mathbf N_{m_p,m_o},
\qquad
\mathbf H_{k,n}\mathbf N_{m_p,m_o}=\mathbf0.
\label{eq:gauge-linear-relations}
\end{equation}
Consequently, the finite-horizon observability matrix $\mathcal O_{k,N}$, which stacks $\mathbf H_{k,n}\mathbf F_{k,n-1}\cdots\mathbf F_{k,0}$ for $n=0,\dots,N$, satisfies $\mathcal O_{k,N}\mathbf N_{m_p,m_o}=\mathbf0$.
That is, the linearized system acquires no information along the six gauge directions.

The point and object initialization Jacobians satisfy
\begin{equation}
\mathbf L_k^q\mathbf N_{m_p,m_o}
=\begin{bmatrix}\mathbf0&\mathbf I\end{bmatrix},
\qquad
\mathbf L_k^o\mathbf N_{m_p,m_o}
=\begin{bmatrix}
\mathbf I&\mathbf0\\
\mathbf0&\mathbf I
\end{bmatrix}.
\label{eq:gauge-initialization-relations}
\end{equation}
After the insertion defined by $\bm\Pi_k^s$, these rows extend the gauge-direction matrix to $\mathbf N_{m_p+1,m_o}$ for a new point feature or $\mathbf N_{m_p,m_o+1}$ for a new object feature.

Therefore, the local propagation, measurement update, and feature initialization preserve the six gauge directions.

Define the gauge-direction matrix of robot \(k\)’s partial state involved in the fusion as $\mathbf N_{kl}^c:=\mathbf E_{kl}\mathbf N_{m_p,m_o}$.
The selected local partial state therefore has error \(\mathbf N_{kl}^c\boldsymbol\xi_g\).
Applying the same global-frame transformation to \eqref{eq:virtual-pose} and \eqref{eq:communicated-state} shows that the communicated estimate has the identical error \(\mathbf N_{kl}^c\boldsymbol\xi_g\).
Using the first-order residual relation \eqref{eq:ci-residual-errors}, the resulting variation of the fusion residual is
\begin{equation}
	\delta\mathbf d_{lk}=
	\left(
	\mathbf E_{kl}\mathbf N_{m_p,m_o}
	-\mathbf N_{kl}^c
	\right)\boldsymbol\xi_g
	=\mathbf0.
	\label{eq:fusion-gauge-relation}
\end{equation}
Thus, the fusion residual is unchanged by a common global rotation or translation and introduces no information along the six gauge directions.
The same argument applies to each communicated residual in the multi-neighbor update.

\subsection{Fusion consistency}

A covariance $\mathbf P$ is first-order consistent for an error $\mathbf e$ if $\mathbb E[\mathbf e\mathbf e^T]\preceq\mathbf P$, where $\preceq$ denotes the Loewner order.

Assume that the local posteriors entering a fusion event satisfy $\mathbb E[\mathbf e_k\mathbf e_k^T]\preceq\mathbf P_k$ and $\mathbb E[\mathbf e_l^{\mathcal S}(\mathbf e_l^{\mathcal S})^T]\preceq\mathbf P_l^{\mathcal S}$.
Within the first-order model in \eqref{eq:compatible-error-mapping}, the zero-mean inter-robot observation noise and the adopted independence assumptions give
\begin{equation}
\begin{split}
\mathbb E[\mathbf e_{l\rightarrow k}\mathbf e_{l\rightarrow k}^T]
&=\mathbb E[\mathbf e_l^{\mathcal S}(\mathbf e_l^{\mathcal S})^T]
+\mathbf D_{l\rightarrow k}\bm\Psi_{lk}
\mathbf D_{l\rightarrow k}^T\\
&\preceq\mathbf P_l^{\mathcal S}
+\mathbf D_{l\rightarrow k}\bm\Psi_{lk}
\mathbf D_{l\rightarrow k}^T
=\mathbf P_{l\rightarrow k}.
\end{split}
\end{equation}
Therefore, \eqref{eq:joint-covariance} is a first-order consistent covariance for the communicated error expressed in coordinates compatible with robot $k$'s local state.

Together with $\mathbf P_k$, this provides the two consistent inputs for partial-state CI, so \eqref{eq:partial-ci-update} preserves first-order consistency without tracking the unknown cross-correlation \cite{ajgl2014partialci,trumpf2025partialci}.
Since $l$ is arbitrary, the same construction applies to every neighbor, and \eqref{eq:multi-neighbor-ci} extends this to multiple neighbors when its weights are nonnegative and sum to one \cite{ajgl2014partialci}.

\section{Experimental Evaluation}

\subsection{Experimental setup}

\begin{figure*}[!htbp]
	\centering
	\begin{minipage}[c]{0.32\textwidth}
		\centering
		\includegraphics[width=\linewidth]{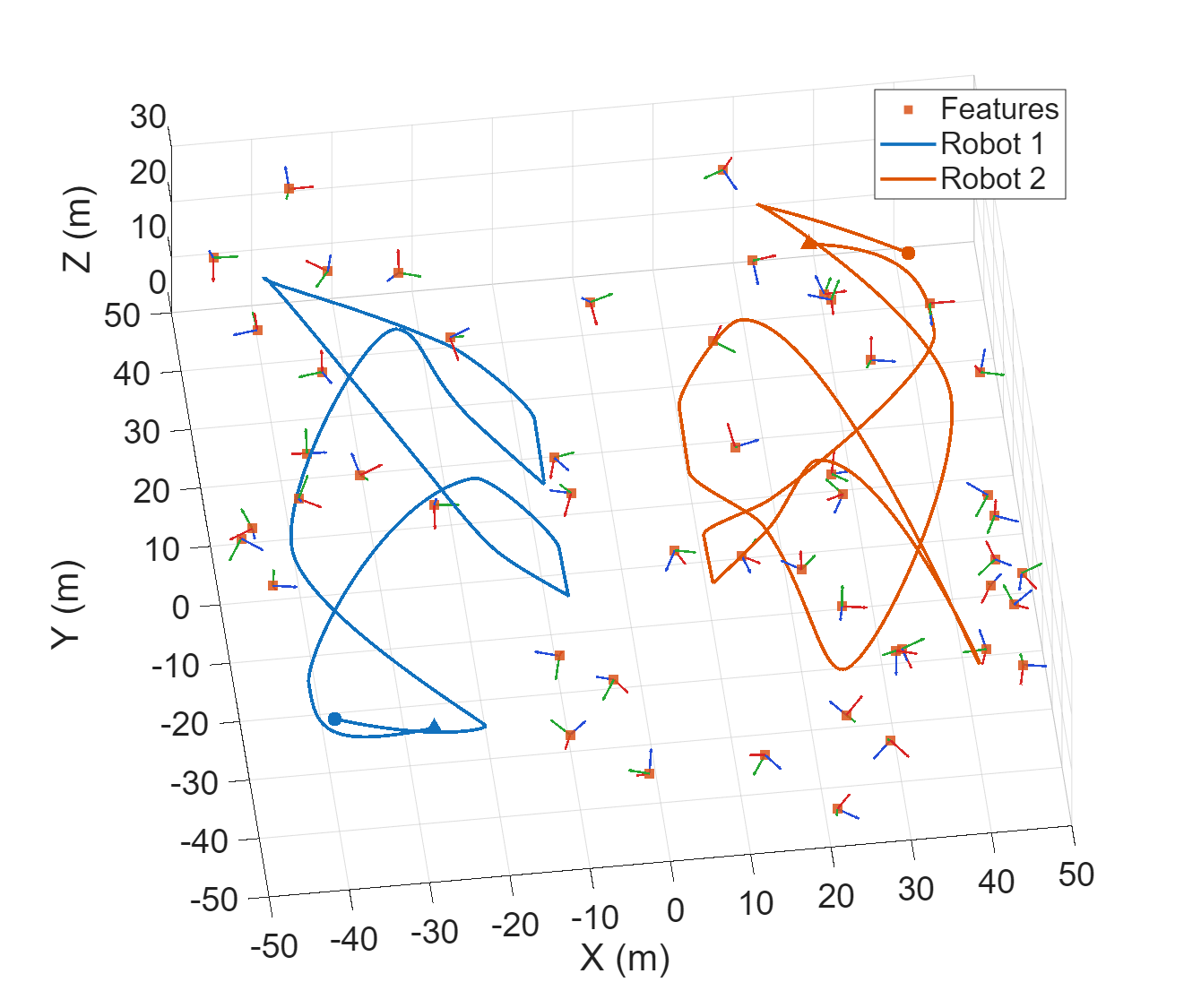}\\[-1mm]
		(a) General scenario
	\end{minipage}
	\hfill
	\begin{minipage}[c]{0.32\textwidth}
		\centering
		\includegraphics[width=\linewidth]{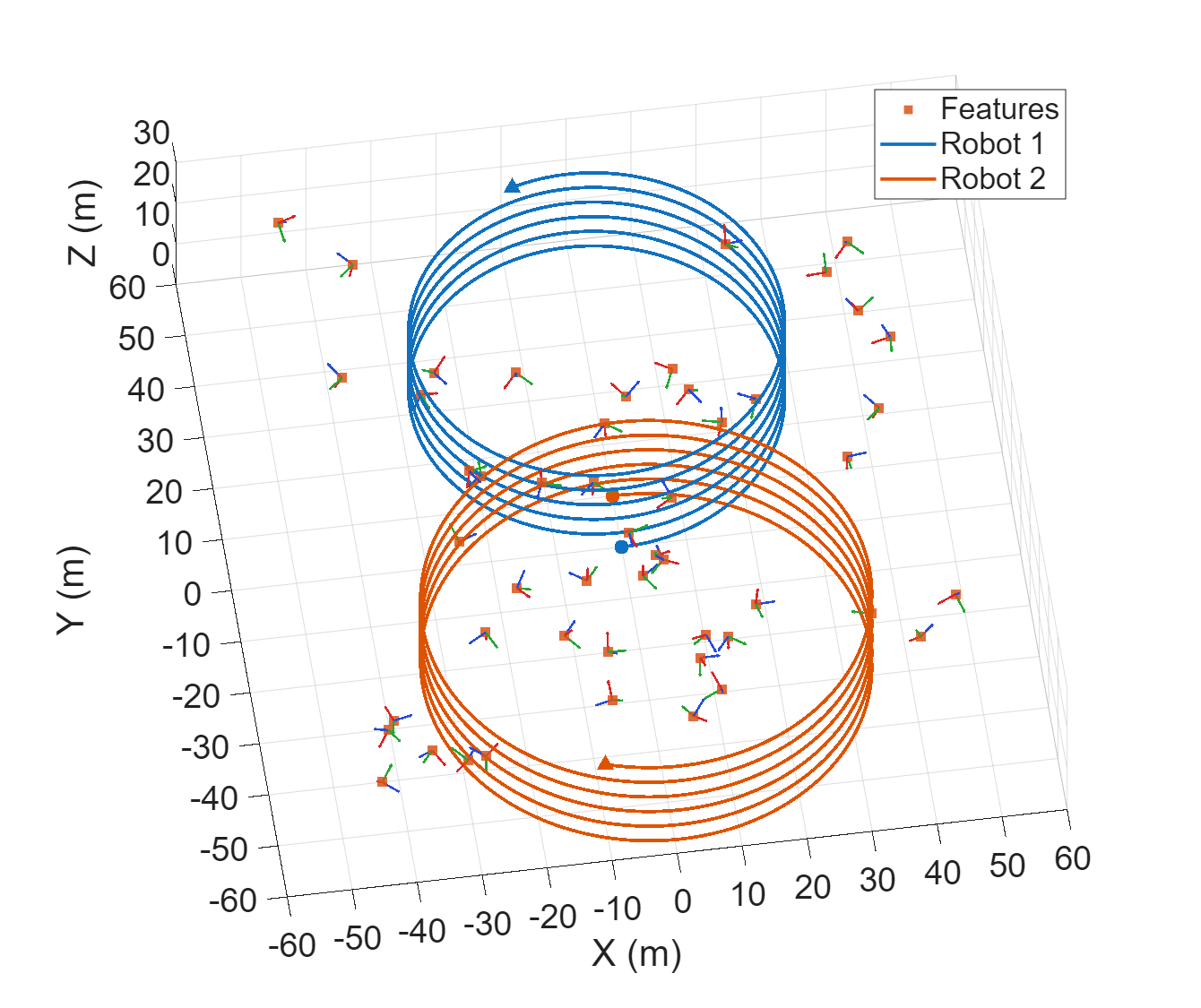}\\[-1mm]
		(b) High-overlap scenario
	\end{minipage}
	\hfill
	\begin{minipage}[c]{0.32\textwidth}
		\centering
		\includegraphics[width=\linewidth]{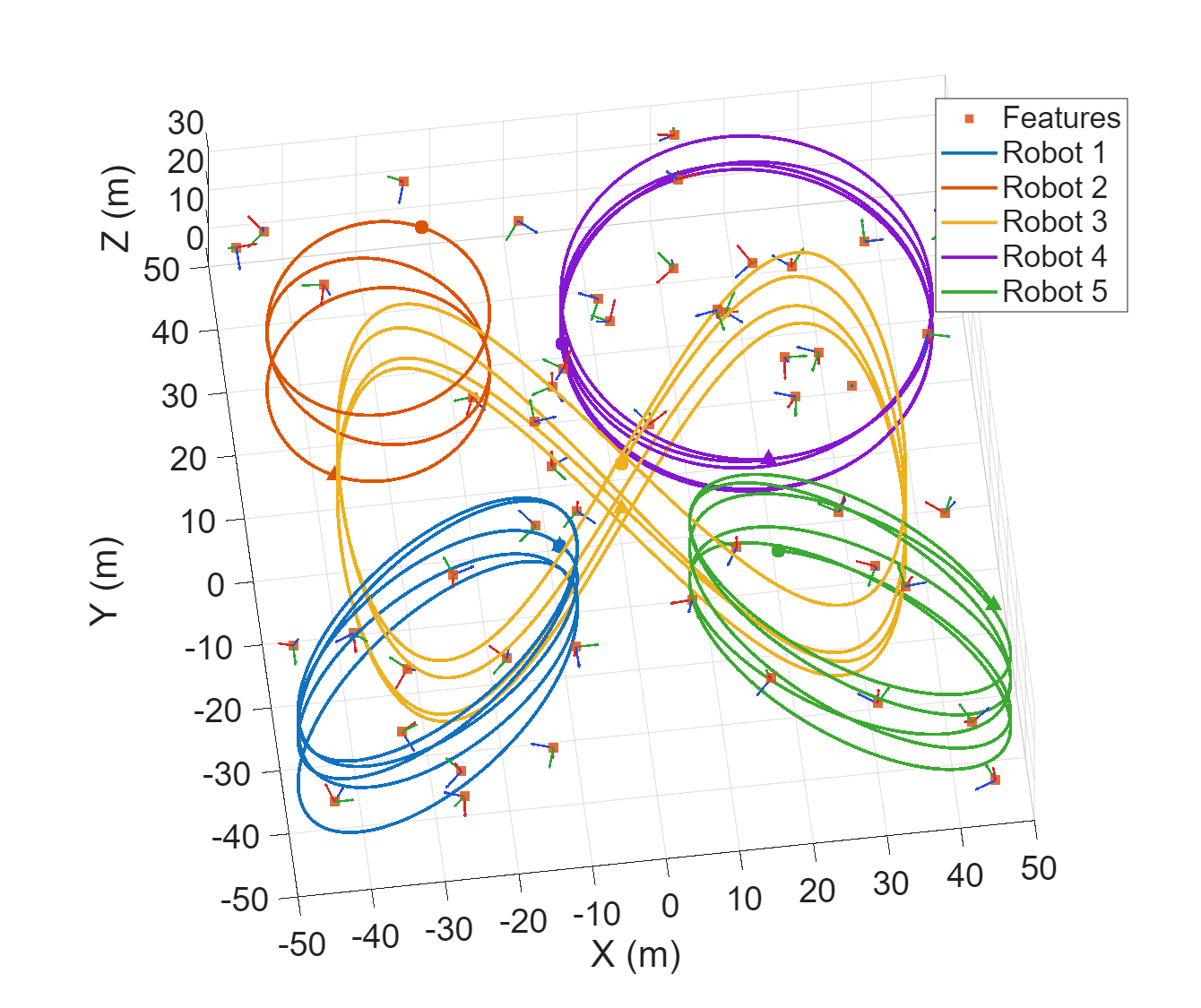}\\[-1mm]
		(c) Five-robot scenario
	\end{minipage}
	\caption{Simulation environments and ground-truth trajectories for the three evaluation scenarios. The same environment and trajectories are used for the point, object, and mixed-map experiments. For visualization, all features are shown as oriented objects with their local coordinate axes.}
	\label{fig:evaluation-scenarios}
\end{figure*}

\begin{figure}[t]
	\centering
	\begin{minipage}[c]{0.38\columnwidth}
		\centering
		\includegraphics[width=\linewidth]{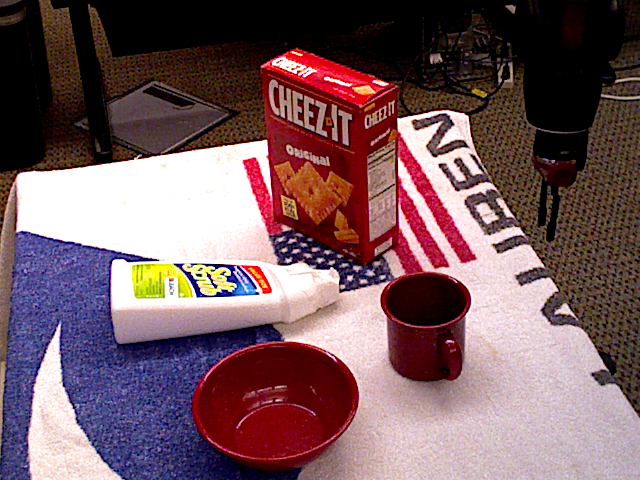}\\[-1mm]
	\end{minipage}
	\begin{minipage}[c]{0.6\columnwidth}
		\centering
		\includegraphics[width=\linewidth]{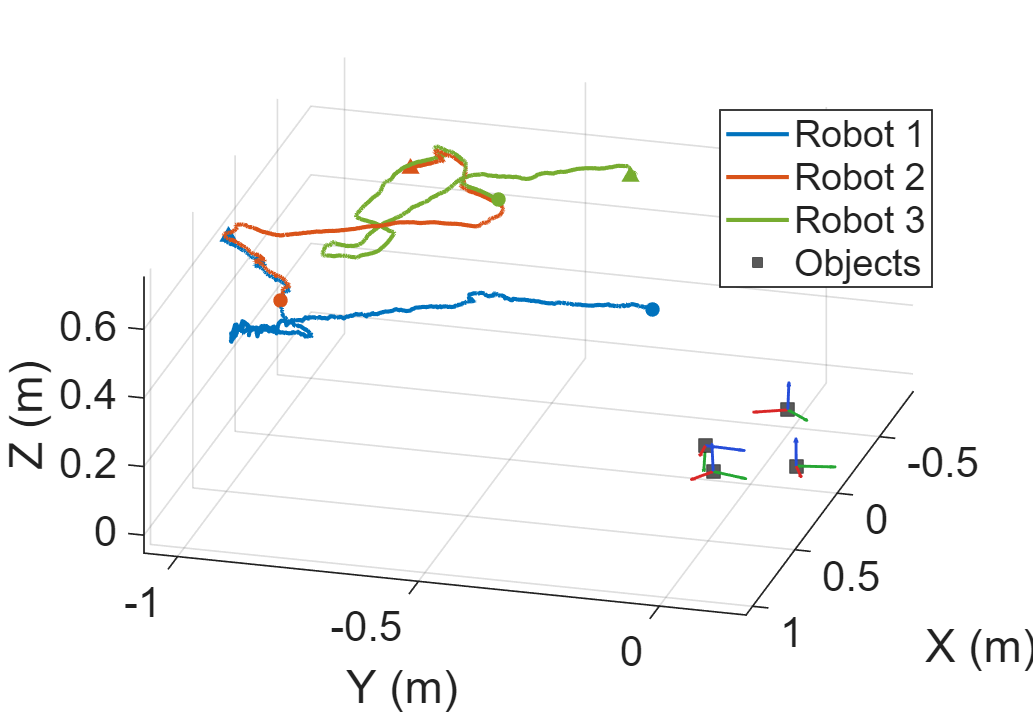}\\[-1mm]
	\end{minipage}
	\caption{YCB-Video dataset-derived evaluation setting. Left: a representative RGB frame. Right: the recorded camera trajectory divided into three overlapping robot trajectories, together with the four object poses.}
	\label{fig:ycbv}
\end{figure}

The three scenarios in Fig.~\ref{fig:evaluation-scenarios} examine complementary operating conditions.
The General scenario uses two robots to provide a controlled comparison across map types under intermittent communication.
The High-overlap scenario tests whether consistency is maintained when frequent communication and extensive shared map information cause repeated information reuse.
The Five-robot scenario provides a more complex and diverse communication setting to test whether the fusion construction remains effective in a multi-neighbor network.
An additional real-world dataset-derived scenario uses Sequence 7 of the YCB-Video dataset \cite{xiang2018posecnn} to evaluate the same backend using recorded camera motion and object geometry, as illustrated in Fig.~\ref{fig:ycbv}.

All three simulation scenarios contain 50 features in a \(100\,\mathrm{m}\times100\,\mathrm{m}\) environment, with each robot following a 2000-time-step trajectory. 
In mixed maps, a fixed randomly selected half of the features are represented as objects and the remainder as points. 
A \(30\,\mathrm{m}\) feature-observation range and a \(20\,\mathrm{m}\) communication range are used throughout.
The common noise covariances are $\bm\Phi=\operatorname{diag}(0.02^2\mathbf I_3,0.5^2\mathbf I_3)$, $\bm\Psi^q=0.5^2\mathbf I_3$, $\bm\Psi^o=0.5^2\mathbf I_6$, and $\bm\Psi_{kl}=0.05^2\mathbf I_6$, with rotational and translational components expressed in radians and meters, respectively.
Two robots within communication range reciprocally exchange messages, using independent inter-robot observations in the two directions.
Both updates at a communication event are computed from the pre-communication posteriors, preventing immediate reuse of information within that event.

To avoid redundant configurations, the High-overlap and Five-robot scenarios use the mixed map as a representative of the unified formulation, which reduces to the point-only case when \(m_o=0\) and the object-only case when \(m_p=0\). 
The object map is retained separately to enable comparison with the object-specific baseline.

\subsubsection{Compared methods}
The comparison includes five methods:
1) Local runs the coupled RI-EKF SLAM estimator without inter-robot fusion.
2) Pose-CI communicates and fuses only the virtual pose of the receiving robot using partial-state CI.
3) Joint-CI (ours) communicates the virtual pose and common features together with their complete covariance in compatible coordinates.
4) DInCIKFc-C is our controlled implementation of the object-level distributed invariant filter DInCIKFc proposed in \cite{li2025objectslam}.
5) DIn-Local uses the local estimator underlying DInCIKFc-C without inter-robot fusion.

DInCIKFc-C retains the state representation and fusion structure of DInCIKFc~\cite{li2025objectslam}, in which the communicated covariance is block-diagonal, assembled from separately constructed robot and object blocks.
For a paired comparison, it uses the same ground truth, observation availability, and physical noise realizations as Joint-CI.
Noise covariances are mapped into the respective tangent coordinates using the corresponding first-order Jacobians.
Historical accumulation of object estimates used in \cite{li2025objectslam} is disabled so that all fusion methods process the same communication events, separating the state and covariance constructions from fusion scheduling. 
Because Joint-CI and DInCIKFc-C also use different local estimators, DIn-Local is included to isolate the effect of inter-robot fusion within DInCIKFc-C.

\subsubsection{Evaluation criteria}
For each scenario and map type, \(100\) Monte Carlo trials are conducted.
Within each trial, all methods use the same ground truth, measurements, noise realizations, visibility sets, and communication events.
The evaluation considers estimation accuracy, statistical consistency, communication payload, and fusion time.

Root-mean-square error (RMSE) measures the accuracy of the robot and the applicable feature components.
The normalized estimation error squared (NEES) and $95\%$ coverage assess statistical consistency.
NEES is computed in each estimator's own error coordinates.
A value close to one indicates that the reported covariance agrees with the observed estimation error.
The \(95\%\) coverage is the fraction of Monte Carlo trials whose quadratic error lies below the $95\%$ chi-square threshold.
A value close to \(0.95\) indicates agreement with the confidence level.

Payload counts the communicated estimate and the independent entries of its symmetric covariance, together with the feature identifiers.
Fusion time includes message construction, CI weight selection, and state update.

\subsection{Two-robot evaluation}
\subsubsection{Setting}

The robots are within communication range for \(10.0\%\) and \(32.0\%\) of the time steps in the General and High-overlap scenarios, with averages of \(8.9\) and \(21.6\) common features per communication event, respectively.

\begin{table*}[!hpbt]
\centering
\caption{Two-robot Monte Carlo comparison. Each paired entry is ordered as robot~1 $\mid$ robot~2.}
\label{tab:main-results}
\begingroup
\renewcommand{\arraystretch}{0.85}
\catcode`\/=\active
\def/{\ensuremath{\,\mid\,}}
\resizebox{\textwidth}{!}{%
\begin{tabular}{c c l c c c c c c c c c c}
\hline
Sc. & Map & Method & $\mathrm{RMSE}_{R}$ & $\mathrm{RMSE}_{x}$ &
$\mathrm{RMSE}_{Rf}$ & $\mathrm{RMSE}_{f}$ &
$\mathrm{NEES}_{r}$ & $\mathrm{NEES}_{p}$ & $C_{95,r}$ & $C_{95,p}$ & Payload & Time \\
\hline
\multirow{11}{*}{G} & \multirow{3}{*}{Point}
& Local          & 7.26/2.57 & 2.06/0.67 & --/-- & 1.99/0.72 & \best{0.94}/\best{1.04} & \best{0.88}/\best{1.07} & 97/90 & \best{96}/\best{92} & --/-- & --/-- \\
& & Pose-CI       & 6.34/2.57 & 1.55/0.67 & --/-- & 1.21/0.72 & 0.73/\best{1.04} & 0.72/\best{1.07} & 100/90 & 99/\best{92} & \best{216}/\best{216} & \best{3.75}/\best{8.11} \\
& & Joint-CI      & \best{6.22}/\best{2.23} & \best{1.49}/\best{0.55} & --/-- & \best{1.10}/\best{0.50} & 0.85/0.88 & 0.79/0.85 & \best{96}/\best{97} & 98/\best{98} & 4792/4792 & 4.72/8.46 \\
\cline{2-13}
& \multirow{5}{*}{Object}
& Local              & 6.27/2.57 & 1.91/0.67 & 10.69/9.50 & 1.94/0.72 & \best{0.96}/\best{1.05} & \best{0.98}/\best{1.08} & \best{96}/89 & \best{95}/\best{92} & --/-- & --/-- \\
& & Pose-CI           & 5.29/2.57 & 1.41/0.67 & 10.21/9.50 & 1.19/0.72 & 0.74/\best{1.05} & 0.78/\best{1.08} & 100/89 & 99/\best{92} & \best{216}/\best{216} & \best{17.65}/45.59 \\
& & Joint-CI          & \best{5.15}/\best{2.22} & \best{1.34}/\best{0.55} & \best{9.69}/\best{8.92} & \best{1.07}/\best{0.50} & 0.86/0.89 & 0.86/0.89 & \best{94}/\best{95} & 98/\best{98} & 15220/15220 & 18.25/\best{41.79} \\
\cline{3-13}
& & DIn-Local         & 6.60/2.88 & 2.03/0.76 & 11.12/9.82 & 2.05/0.82 & \best{1.11}/\best{1.32} & \best{1.06}/\best{1.24} & \best{92}/\best{79} & \best{93}/\best{88} & --/-- & --/-- \\
& & DInCIKFc-C        & \best{5.43}/\best{2.53} & \best{1.45}/\best{0.63} & \best{10.18}/\best{9.14} & \best{1.17}/\best{0.58} & 1.24/1.34 & 1.48/1.25 & 74/\best{79} & 58/80 & 2167/2167 & 17.70/42.65 \\
\cline{2-13}
& \multirow{3}{*}{Mixed}
& Local          & 6.29/2.57 & 1.92/0.67 & 10.07/10.10 & 1.94/0.72 & \best{0.95}/\best{1.04} & \best{0.93}/\best{1.08} & 96/91 & \best{96}/\best{92} & --/-- & --/-- \\
& & Pose-CI       & 5.31/2.57 & 1.41/0.67 & 9.56/10.10 & 1.18/0.72 & 0.74/\best{1.04} & 0.75/\best{1.08} & 100/91 & 99/\best{92} & \best{216}/\best{216} & \best{8.81}/20.53 \\
& & Joint-CI      & \best{5.17}/\best{2.22} & \best{1.35}/\best{0.55} & \best{9.17}/\best{9.21} & \best{1.07}/\best{0.50} & 0.85/0.88 & 0.83/0.87 & \best{95}/\best{97} & 98/\best{98} & 8883/8883 & 9.48/\best{19.41} \\
\hline
\noalign{\vskip 1.6pt}
\hline
\multirow{8}{*}{H}
& \multirow{5}{*}{Object}
& Local              & 6.36/2.93 & 1.51/0.87 & 12.00/11.68 & 0.65/0.66 & \best{1.02}/\best{1.00} & \best{1.06}/\best{1.04} & \best{95}/\best{94} & \best{94}/\best{96} & --/-- & --/-- \\
& & Pose-CI           & 6.36/2.93 & 1.51/0.87 & 12.00/11.68 & 0.65/0.66 & \best{1.02}/\best{1.00} & \best{1.06}/\best{1.04} & \best{95}/\best{94} & \best{94}/\best{96} & \best{216}/\best{216} & \best{35.08}/57.61 \\
& & Joint-CI          & \best{6.32}/\best{2.56} & \best{1.49}/\best{0.62} & \best{11.49}/\best{10.41} & \best{0.61}/\best{0.41} & 0.89/0.79 & 0.92/0.86 & 99/98 & 97/98 & 76917/76917 & 35.89/58.54 \\
\cline{3-13}
& & DIn-Local         & 6.56/3.29 & 1.57/0.97 & 12.71/12.12 & 0.69/0.73 & \best{1.27}/\best{1.28} & \best{1.70}/\best{1.26} & \best{92}/\best{84} & \best{83}/\best{87} & --/-- & --/-- \\
& & DInCIKFc-C        & \best{6.51}/\best{2.92} & \best{1.54}/\best{0.72} & \best{11.86}/\best{10.84} & \best{0.63}/\best{0.48} & 1.97/2.08 & 5.04/4.83 & 81/44 & 10/13 & 4960/4960 & 36.01/\best{57.19} \\
\cline{2-13}
& \multirow{3}{*}{Mixed}
& Local          & 6.70/2.93 & 1.61/0.87 & 11.99/10.87 & 0.67/0.66 & \best{1.01}/\best{0.99} & \best{1.06}/\best{1.02} & \best{95}/\best{95} & \best{95}/\best{97} & --/-- & --/-- \\
& & Pose-CI       & 6.70/2.93 & 1.61/0.87 & 11.99/10.87 & 0.67/0.66 & \best{1.01}/\best{0.99} & \best{1.06}/\best{1.02} & \best{95}/\best{95} & \best{95}/\best{97} & \best{216}/\best{216} & \best{16.44}/\best{28.27} \\
& & Joint-CI      & \best{6.67}/\best{2.56} & \best{1.59}/\best{0.62} & \best{11.52}/\best{9.84} & \best{0.63}/\best{0.41} & 0.89/0.78 & 0.92/0.83 & 99/98 & 97/98 & 48728/48728 & 17.55/29.66 \\
\hline
\end{tabular}}
\endgroup
\begin{tablenotes}
	\item \textit{Metrics and units:} $\mathrm{RMSE}_{R}$/$\mathrm{RMSE}_{Rf}$, robot/feature rotation RMSE ($10^{-2}\,\mathrm{rad}$); $\mathrm{RMSE}_{x}$/$\mathrm{RMSE}_{f}$, robot/feature position RMSE ($\mathrm{m}$); $\mathrm{NEES}_{r}$, robot pose NEES; $\mathrm{NEES}_{p}$, mean normalized marginal feature NEES (3-D point position or 6-D object pose); $C_{95,r}$/$C_{95,p}$, final-step robot-pose/mean marginal feature coverage (\%); Payload, bytes; Time, fusion time (ms), both per communication direction and event.
	\item \textit{Notation:} G, General scenario; H, High-overlap scenario. -- denotes an inapplicable entry. Bold red marks the lowest RMSE, NEES closest to one, and coverage closest to $95\%$ within each local-estimator group. For payload and time, it marks the lowest value among all fusion methods.
\end{tablenotes}
\end{table*}

\subsubsection{Results}
Table~\ref{tab:main-results} summarizes the two-robot results. 
Regarding accuracy, the RMSE results show that, in the General scenario, Pose-CI improves robot~1 for all three map types, whereas the more accurate estimate of robot~2 remains unchanged.
The optimized CI weights of Pose-CI show that robot~2 consistently assigns unit weight to its local posterior, so the communicated estimate from robot~1 is rejected.
In the High-overlap scenario, the optimized CI weight on the local posterior equals one in both communication directions, so Pose-CI produces no effective fusion.
This indicates that the local maps already contain substantial common information, and the pose-only message provides insufficient new information to reduce the minimum-trace objective.
In both scenarios, Joint-CI reduces the RMSEs of both robots and achieves the lowest RMSEs, demonstrating the benefit of communicating the common features together with their joint covariance. 
For object maps, DInCIKFc-C also improves both robots over DIn-Local.
When normalized by their respective local baselines, DInCIKFc-C and Joint-CI achieve similar relative RMSE reductions.

\begin{figure}[!hpbt]
	\centering
	\includegraphics[width=\linewidth]{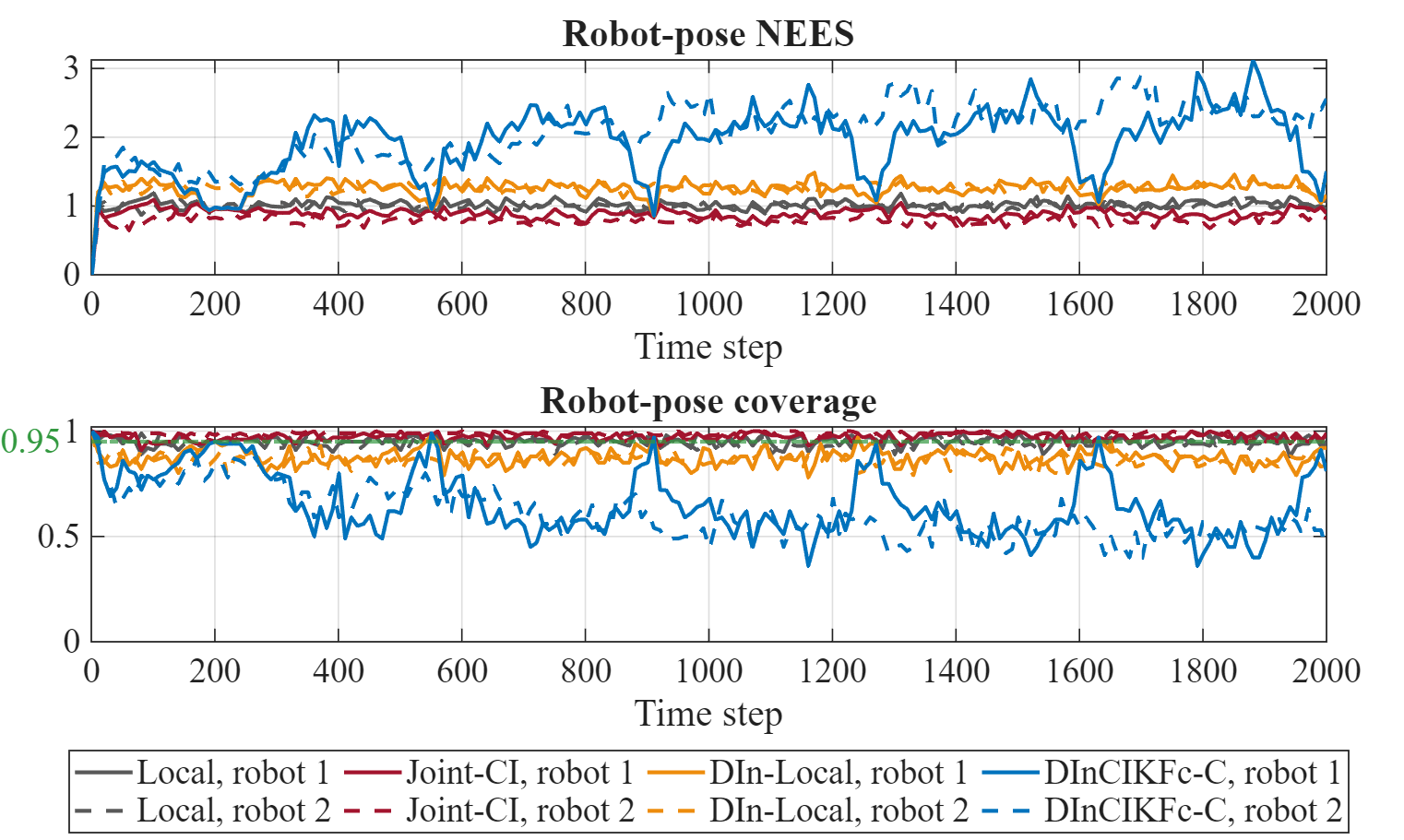}
	\caption{Robot-pose NEES and $95\%$ robot-pose coverage in the High-overlap object-map scenario.}
	\label{fig:main-consistency}
\end{figure}

Joint-CI yields NEES values slightly below one and coverage at or above the nominal level in most cases, indicating consistent and mildly conservative uncertainty estimates. 
This conservatism is expected for CI, which bounds rather than models the unknown cross-correlation.
When Pose-CI performs effective fusion, its estimates are generally more conservative.
In contrast, DInCIKFc-C produces larger NEES and lower coverage, especially in the High-overlap scenario, indicating overconfident uncertainty estimates.
Fig.~\ref{fig:main-consistency} further presents the time-resolved robot-pose NEES and \(95\%\) coverage in the High-overlap object-map scenario, with Pose-CI omitted because no effective fusion occurs.
Throughout the trajectory, both local estimators and Joint-CI remain stable without progressive consistency degradation.
In contrast, DInCIKFc-C shows increasing NEES and decreasing coverage under frequent and repeated information exchange.
A plausible cause is its block-diagonal communicated covariance.
Although the per-object invariant errors avoid re-expressing the object errors, the virtual pose still inherits the neighboring robot's pose error, which is correlated with the common object estimates.
A covariance assembled without these nonzero cross-covariance blocks is not guaranteed to upper-bound the actual error covariance.

Joint-CI achieves this consistency by communicating the complete covariance of the correlated robot--feature partial state, whose payload grows quadratically with the number of common features but is independent of the local map size. 
Since any principal submatrix of a consistent covariance remains consistent (the Loewner order is preserved under congruence), any subset of common features can be selected while preserving first-order covariance consistency, with Pose-CI as the special case of the empty subset. 
This enables payload reduction through feature selection.
The fusion time of Joint-CI remains comparable to that of the baselines.

\subsection{Multi-robot network evaluation}

\subsubsection{Setting}
In the Five-robot scenario, at least one communication edge is active during \(49.3\%\) of the time steps, with an average of \(20.0\) common features per edge event. 
The largest connected component contains at least three robots during \(3.25\%\) of the time steps.
Joint-CI directly applies the multi-estimate partial-state CI update in \eqref{eq:multi-neighbor-ci}.
DInCIKFc-C first combines neighboring messages using FastCI and then fuses the resulting message with the local posterior using \eqref{eq:partial-ci-update}, following the two-stage structure in \cite{li2025objectslam}.
For a controlled comparison, Joint-FastCI applies the same two-stage structure to the coordinate-compatible joint messages.

Following \cite{li2025objectslam}, the dataset-derived experiment divides the recorded camera trajectory into three 672-step segments, one per robot, with a \(30\%\) overlap.
The robots communicate reciprocally every ten time steps over a fixed chain graph \(1\leftrightarrow2\leftrightarrow3\), where robot~2 fuses messages from both neighbors at every communication event.
Odometry, object, and inter-robot observations are generated from the recorded ground-truth geometry and independently perturbed in each Monte Carlo trial, while the recorded frame-wise object availability is preserved.
The noise covariances are
\(\bm\Phi=\operatorname{diag}(0.01^2\mathbf I_3,0.01^2\mathbf I_3)\),
\(\bm\Psi^o=\operatorname{diag}(0.05^2\mathbf I_3,0.01^2\mathbf I_3)\), and
\(\bm\Psi_{kl}=\operatorname{diag}(0.01^2\mathbf I_3,0.01^2\mathbf I_3)\).

\begin{table*}[!hpbt]
	\centering
	\caption{Five-robot simulation and YCB-Video dataset-derived Monte Carlo comparison.}
	\label{tab:multi-results1}
	\resizebox{\textwidth}{!}{%
		\begin{tabular}{c c l c c c c c c c c c c}
			\hline
			Sc. & Map & Method & $\mathrm{RMSE}_{R}$ & $\mathrm{RMSE}_{x}$ &
			$\mathrm{RMSE}_{Rf}$ & $\mathrm{RMSE}_{f}$ &
			$\mathrm{NEES}_{r}$ & $\mathrm{NEES}_{p}$ & $C_{95,r}$ & $C_{95,p}$ & Payload & Time \\
			\hline
			\multirow{9}{*}{Five-R} & \multirow{6}{*}{Object}
			& Local        & $2.46\pm0.36$ & $54.45\pm4.51$ & $9.81\pm1.98$ & $44.26\pm6.62$ & \best{1.00} & \best{1.06} & \best{94} & \best{93} & -- & -- \\
			& & Pose-CI      & $3.50\pm0.79$ & $88.98\pm31.16$ & $12.02\pm3.06$ & $82.47\pm37.31$ & 0.81 & 0.87 & 99 & 99 & \best{231} & 51.12 \\
			& & Joint-CI     & $\bestmath{2.27}\pm0.35$ & $48.78\pm4.77$ & $9.16\pm1.20$ & $34.00\pm4.39$ & 0.84 & 0.85 & 98 & 98 & 71134 & \best{44.27} \\
			& & Joint-FastCI & $\bestmath{2.27}\pm0.34$ & $\bestmath{48.49}\pm4.84$ & $\bestmath{8.97}\pm1.15$ & $\bestmath{33.85}\pm4.26$ & 0.85 & 0.90 & 98 & \best{97} & 71134 & 44.60 \\
			\cline{3-13}
			& & DIn-Local     & $2.76\pm0.38$ & $61.35\pm5.00$ & $10.14\pm2.03$ & $49.55\pm8.02$ & \best{1.27} & \best{1.20} & \best{86} & \best{90} & -- & -- \\
			& & DInCIKFc-C   & $\bestmath{2.43}\pm0.41$ & $\bestmath{50.72}\pm7.09$ & $\bestmath{8.57}\pm1.10$ & $\bestmath{31.16}\pm2.62$ & 1.56 & 2.45 & 69 & 33 & {4803} & 51.40 \\
			\cline{2-13}
			& \multirow{3}{*}{Mixed}
			& Local        & $2.46\pm0.37$ & $54.58\pm4.70$ & $9.42\pm1.41$ & $44.30\pm6.71$ & \best{0.99} & \best{1.05} & \best{95} & \best{93} & -- & -- \\
			& & Pose-CI      & $3.64\pm0.89$ & $92.69\pm34.34$ & $11.66\pm2.27$ & $86.53\pm40.66$ & 0.81 & 0.85 & 99 & 99 & \best{231} & \best{20.16} \\
			& & Joint-CI     & $\bestmath{2.28}\pm0.35$ & $\bestmath{49.12}\pm4.74$ & $\bestmath{9.09}\pm1.25$ & $\bestmath{34.33}\pm4.40$ & 0.83 & 0.85 & 99 & \best{97} & 42237 & 21.02 \\
			\hline
			\multirow{5}{*}{YCB-V} & \multirow{5}{*}{Object}
			& Local      & $2.85\pm0.04$ & $2.62\pm0.15$ & $2.09\pm0.03$ & $0.81\pm0.03$ & \best{1.00} & \best{1.00} & 94 & \best{95} & -- & -- \\
			& & Pose-CI    & $2.91\pm0.15$ & $2.69\pm0.15$ & $2.43\pm0.61$ & $1.10\pm0.49$ & 0.84 & 0.82 & \best{95} & 97 & \best{288} & \best{1.39} \\
			& & Joint-CI   & $\bestmath{2.67}\pm0.12$ & $\bestmath{2.47}\pm0.28$ & $\bestmath{1.87}\pm0.18$ & $\bestmath{0.73}\pm0.10$ & 0.89 & 0.81 & 97 & 98 & 5301 & 1.60 \\
			\cline{3-13}
			& & DIn-Local  & $2.85\pm0.05$ & $2.62\pm0.15$ & $2.09\pm0.03$ & $0.81\pm0.03$ & \best{1.00} & \best{1.00} & \best{95} & \best{95} & -- & -- \\
			& & DInCIKFc-C & $\bestmath{2.44}\pm0.02$ & $\bestmath{2.24}\pm0.17$ & $\bestmath{1.53}\pm0.01$ & $\bestmath{0.55}\pm0.01$ & 1.34 & 6.52 & 84 & 6 & 1461 & 1.55 \\
			\hline
	\end{tabular}}
	\begin{tablenotes}
		\item Five-R and YCB-V denote the Five-robot and YCB-Video scenarios. Position RMSEs are in centimeters; other units and highlighting follow Table~I. RMSEs are reported as mean \(\pm\) standard deviation across robots. The remaining entries are equally weighted averages across robots.
	\end{tablenotes}
\end{table*}

\subsubsection{Results}
Table~\ref{tab:multi-results1} evaluates the proposed construction in multi-neighbor settings. 
Across the Five-robot and dataset-derived experiments, Joint-CI reduces the average RMSE relative to Local while maintaining consistent, mildly conservative uncertainty estimates. 
Pose-CI increases the average RMSE while its uncertainty remains conservative, indicating a loss of information rather than overconfidence.
A plausible cause is that each pose-only fusion discounts the entire local posterior, including the map, whereas new information enters only the pose.
DInCIKFc-C achieves larger RMSE reductions relative to DIn-Local, but its larger NEES and lower coverage indicate overconfident uncertainty estimates. 
In the Five-robot experiment, Joint-FastCI closely matches direct Joint-CI, showing that the compatible joint messages also support two-stage neighbor aggregation. 

For the mixed map, Joint-CI exhibits the same overall behavior, supporting the same construction when point and object features are represented in a unified state.
The main cost of Joint-CI is its larger communication payload.

\section{Conclusion}

This paper presented Joint-CI for distributed RI-EKF SLAM with point, object, and mixed feature maps.
A direct error mapping constructs a correlated robot--feature estimate with a covariance compatible with the receiving filter, after which standard partial-state CI can be applied without tracking inter-robot correlations.
The analysis establishes preservation of the gauge directions and first-order covariance consistency of communication and fusion given consistent local inputs.
Across the simulations and the real-world dataset-based experiment, Joint-CI reduces RMSE relative to the corresponding local RI-EKF and achieves lower RMSE than Pose-CI while maintaining consistent, mildly conservative uncertainty under repeated information exchange.
This improvement requires transmitting the complete joint covariance and therefore incurs a larger communication payload.

Future work will investigate communication-aware fusion scheduling and feature selection, as well as extensions to unknown initial relative robot poses with uncertain transformations between local maps.

\bibliographystyle{IEEEtran}

\bibliography{Reference}

\end{document}